\documentstyle[elsart12,named,lingmacros,times]{article} 

\title{Centering, Anaphora Resolution, and Discourse Structure\\
}

\author{ Marilyn A. Walker\\  ATT Labs Research  \\  180 Park Ave. \\
Florham Park, N.J. 07932 \\ {\tt walker@research.att.com}}

\input{psfig} \date{}

\begin{document}           

\bibliographystyle{named} 

\maketitle 

\begin{abstract}

Centering was formulated as a model of the relationship between
attentional state, the form of referring expressions, and the
coherence of an utterance {\it within a discourse segment} (Grosz,
Joshi and Weinstein, 1986; Grosz, Joshi and Weinstein, 1995). In this
chapter, I argue that the restriction of centering to operating within
a discourse segment should be abandoned in order to integrate
centering with a model of global discourse structure. The
within-segment restriction causes three problems. The first problem is
that centers are often continued over discourse segment boundaries
with pronominal referring expressions whose form is identical to those
that occur within a discourse segment.  The second problem is that
recent work has shown that listeners perceive segment boundaries at
various levels of granularity. If centering models a universal
processing phenomenon, it is implausible that each listener is using a
different centering algorithm.The third issue is that even for
utterances within a discourse segment, there are strong contrasts
between utterances whose adjacent utterance within a segment is
hierarchically recent and those whose adjacent utterance within a
segment is linearly recent.  This chapter argues that these problems
can be eliminated by replacing Grosz and Sidner's stack model of
attentional state with an alternate model, the cache model. I show how
the cache model is easily integrated with the centering algorithm, and
provide several types of data from naturally occurring discourses that
support the proposed integrated model.  Future work should provide
additional support for these claims with an examination of a larger
corpus of naturally occurring discourses.

\end{abstract}


\section{Introduction}

\label{intro-sec}

Centering is formulated as {\it a theory that relates focus of
attention, choice of referring expression, and perceived
coherence of utterances, within a discourse segment}
\cite{GJW95}, p. 204. In this chapter, I argue that the
restriction of centering to utterances within the same discourse
segment poses three problems for the theory that can be
eliminated by abandoning this restriction, and  integrating
centering with the cache model of attentional state proposed in
\cite{Walker96b}.

The first problem is that centers are often continued over
discourse segment boundaries with pronominal referring
expressions whose form is identical to those that occur within a
discourse segment.  For example, consider discourse A, a 
naturally occurring discourse excerpt from the {\it Pear
Stories} \cite{Chafe80,Passonneau95}:

\eenumsentence[A]
{\item[] (29) and  he$_{i}$ 's going to take  a pear
or two,  and then.. go on his way \\
(30) um but {\it the little boy$_i$} comes, \\ 
(31) and uh he$_i$ doesn't want just a pear,\\   
(32) {\it he$_i$} wants a whole basket.\\  
(33)  So {\it he$_i$} puts the bicycle down, \\   
(34)  and he$_i$ ... }

In an experiment where naive subjects coded discourses for
segment structure \cite{Passonneau95}, a majority of subjects
placed a discourse segment boundary between utterances (32) and
(33).  If utterance (32) and (33) were subjected to a centering
analysis (cf.  Walker, Joshi and Prince, this volume), (33)
realizes a {\sc continue} transition, indicating that utterance
(33) is highly coherent in the context of utterance (32). It
seems implausible that a different process than centering would
be required to explain the relationship between utterances (32)
and (33), simply because these utterances span a discourse
segment boundary.

 The second problem is that listeners perceive segment boundaries at
various levels of granularity
\cite{PassonneauLitman93,Hearst94,FlammiaZue95b,HirschbergNakatani96},
and some segment boundaries are 'fuzzy' \cite{PassonneauLitman96}. For
example in discourse A above, 5 out of 7 subjects placed a segment
boundary between utterances 29 and 30, while 4 out of 7 subjects
placed a segment boundary between utterances 32 and 33
\cite{Passonneau95}.  If centering models a universal processing
phenomenon, it is implausible that the subjects that place a segment
boundary in these locations {\bf don't} use centering to process the
referring expressions in the discourse, while the subjects who didn't
place a segment boundary {\bf do} use centering for discourse
processing.

\begin{figure}[htb]
\centerline{\psfig{figure=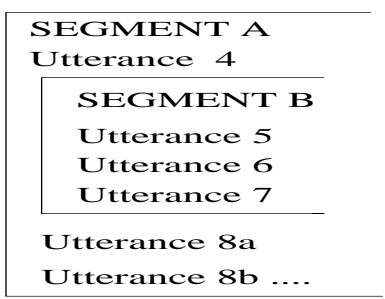,height=1.5in,width=2.0in}}
\caption{The discourse structure of Dialogue B.}
\label{diala-fig} \end{figure}

The third issue is that even for utterances within a discourse
segment, there are strong contrasts between utterances whose
adjacent utterance within a segment is hierarchically recent
and those whose adjacent utterance within a segment is linearly
recent.  
Briefly, an utterance $U_i$ is linearly recent for a subsequent
utterance $U_{i+j}$ if
$U_i$ occurred within the last few utterances. An utterance
$U_i$ is hierarchically recent for a subsequent utterance
$U_{i+j}$ if  $U_{i+j}$ can become adjacent to 
$U_i$ as a result of Grosz and Sidner's stack mechanism
\cite{GS86,Walker96b}. For example consider the contrast between
discourses B and C below, where C is a constructed variation of
B  \cite{PHW82}:\footnote{This dialogue is from a corpus of
naturally occurring financial advice dialogues that were
originally taped from a live radio broadcast and transcribed by
Martha Pollack and Julia Hirschberg. I am grateful to Julia
Hirschberg for providing me with audio tapes of these dialogues.}

\eenumsentence[B]{ \item[] (4) C: Ok Harry, I have a problem
that uh my - with today's economy {\it my daughter is working},
\\ (5) H: I missed your name. \\ (6) C: Hank. \\ (7) H: Go ahead
Hank \\ (8a) C: {\it as well as her uh husband}. \\ (8b) They
have a child. \\ (8c) and they bring the child to us every day
for babysitting. \label{hank-dial} }

\eenumsentence[C]{ \item[] (4) C: Ok Harry, I have a problem
that uh my - with today's economy  {\it my daughter is working},
\\  (5) H: I missed your name. \\  (6) C: Hank. \\ H: I'm sorry,
I can't hear you. \\ C: Hank. \\ H: Is that H A N K? \\  C: Yes.
\\  (7) H: Go ahead Hank. \\  (8a) C:  {\it as well as her uh
husband}. \\  (8b) They have a child. \\ (8c) and they bring the
child to us every day for babysitting. 
\label{hank-construct-dial} }

The structure of Dialogue B is represented schematically in
Figure \ref{diala-fig}.  In utterance 5 of dialogue B, the talk
show host, H , interrupts the caller C to ask for his name. In
utterance 8a, the caller C continues the problem statement that
he began with utterance 4 as though utterance 4 had just been
said, and so utterance 8a is part of the same discourse segment
as utterance 4.  The structure of Dialogue C is identical to
that of B.

But if utterance 8a is in the same segment as utterance 4 in
both dialogue B and C, there is an unexpected difference in the
coherence of the utterance. The anaphoric referring expression,
{\it her husband} is clearly more difficult to interpret in C.
Thus hierarchical recency, as operationalized by the stack
model, does not predict when previous centers are accessible.

I will argue that it is possible to integrate centering with a model
of global discourse structure and simultaneously address these
problems by replacing Grosz and Sidner's stack model of global focus
with the cache model of attention state proposed in
\cite{Walker96b}.\footnote{The cache model is an extension of the AWM
model in \cite{Walker93c,Walker94a,JordanWalker96}.} In the resulting
integrated model:

\begin{enumerate}

\item Centers are elements of the cache and the cache model
mediates the accessibility of centers.

\item Centers are carried over segment boundaries by default.

\item Processing difficulties are predicted for the
interpretation of centers whose co-specifiers are not linearly
recent, as in the case of Dialogue C.

\item Granularity of discourse segmentation has no effect on the
model.

\end{enumerate}

The structure of the chapter is as follows.  Section
\ref{cache-sec} presents the proposed cache model, and section
\ref{alg-sec} defines a version of the centering algorithm
\cite{BFP87} that is integrated with the cache model. Then,
three types of evidence are used to support the proposed
integrated model. First, section \ref{fs-sec} presents evidence
that the cache model can handle `focus pops', a phenomenon that
was believed to provide strong support for Grosz and Sidner's
stack model.  Then section \ref{now-sec} discuss quantitative
evidence showing that centers are frequently carried over
segment boundaries. Next, section \ref{seg-sec} discuss a number
of naturally occurring examples that illustrate that the form in
which centers are realized across discourse segment boundaries
is not determined by boundary type. Finally, section
\ref{conc-sec} summarizes the discussion and outlines future
work.

\section{The Cache Model of Attentional State} \label{cache-sec}

A cache is an easily accessible temporary location used for
storing information that is currently being used by a
computational procedure \cite{Stone87}.  The fundamental idea of
the cache model is that the functioning of the cache when
processing discourse is analogous to that of a cache when
executing a program on a computer. Just as discourses may be
structured into goals and subgoals which contribute to achieving
the purpose of the discourse, a computer program is
hierarchically structured into routines and subroutines which
contribute to completing the routine. Thus a cache can be used
to model attentional state when intentions are hierarchically
structured, just as a cache can be used for processing the
references and operations of a hierarchically structured
program.

In the cache model there are two types of memory: {\sc main
memory} represents long-term memory and the {\sc cache}
represents working memory \cite{Baddeley86}. Main memory is 
much larger than the cache, but is slower to access
\cite{Hintzman88,GillundSchiffrin84}.  The cache is a limited
capacity, almost instantaneously accessible, memory store.  The
size of the cache is a working assumption based on the findings
of previous work \cite{Kintsch88,Miller56,alshawi87}:

\begin{quote} {\sc cache size assumption}: The cache is limited
to 2 or 3 sentences, or approximately 7 propositions. \end{quote}

Given  a particular cache size assumption, the definition of
linear recency, discussed
briefly above, can be made more precise, by setting the number
of linearly adjacent utterances
to be equal to the cache size parameter.

\begin{quote} An utterance U$_i$ is linearly recent for
utterance U$_j$ when it occurred within  the past three 
linearly adjacent utterances. \end{quote}

There are three operations involving the cache and main memory.
Items in the cache can be preferentially {\sc retained} and
items in main memory can be {\sc retrieved} to the cache. Items
in the cache can also be {\sc stored} to main memory.  When new
items are retrieved from main memory to the cache, or enter the
cache directly due to events in the world, other items may be
displaced to main memory, because the cache has limited
capacity.  

The determination of which items to displace is handled by a
{\sc cache replacement policy}. In the cache model, the cache
replacement policy is a working assumption, based on previous
work on the effects of distance on anaphoric processing
\cite{CS79,Hobbs76a,HS76} {\it inter alia}:

\begin{quote} {\sc cache replacement policy  assumption}: The
least recently accessed  items in the  cache are displaced to
main memory, with the exception of those items preferentially
retained. \end{quote}

The cache model includes specific assumptions about processing.
Discourse processes execute on elements that are in the cache. 
All of the premises for an inference must be simultaneously in
the cache for the inference to be made
\cite{McKoonRatcliff92,Goldman86}. If a discourse relation is to
be inferred between two separate segments, a representation of
both segments must be simultaneously in the cache
\cite{FHM90,Walker93c}. The cospecifier of an anaphor must  be
in the cache for automatic interpretation or be strategically
retrieved to the cache in order to interpret the anaphor
\cite{TM82,GMR92}.  Thus what is contained in the cache at any
one time is a {\sc working set} consisting of discourse entities
such as entities, properties and relations that are currently
being used for some process.

In the cache model, centers are a subset of entities in the
cache, and the contents of the cache change incrementally as
discourse is processed utterance by utterance, so by  default 
centers are carried over from one segment to another The cache
model is easily integrated with the centering rules and
constraints by simply assuming that the Cf list for an utterance
is a subset of the entities in the cache, and that the centering
rules and constraints apply as usual, with the ordering of the
Cf list providing an additional finer level of salience ordering
for entities within the cache.  .

The cache model maintains Grosz and Sidner's distinction between
intentional structure and attentional state. This distinction is
critical. However the cache model does not posit that
attentional state is isomorphic to intentional structure. For
example, when a new intention is recognized that is subordinate
to the current intention, new entities may be created in the
cache or be retrieved to the cache from main memory
\cite{RatcliffMcKoon88}, however old entities currently in the
cache will remain until they are displaced. Thus centers from
the previous intention are carried over by default until they
are displaced. When a new intention that is subordinate to a
prior intention is recognized, entities related to the prior
intention must be retrieved to the cache, unless they were not
displaced by the intervening discourse. In other words, the
cache model casts attentional state in discourse processing as a
{\bf gradient} phenomenon, and predicts a looser coupling of
intentional structure and attentional state. A change of
intention affects what is in the cache, but the contents of the
cache change incrementally, instead of changing instantaneously
with one stack operation as they do with in stack model.

The cache model provides a natural explanation for the
difference in the coherence between dialogue B and dialogue C.
The {\sc cache size assumption} in the cache model predicts that
processing the longer interruption in C uses all of the cache
capacity; thus returning to the prior discussion requires a
retrieval from main memory. The success of this retrieval
depends on two requirements: (1) the speaker must provide an
adequate retrieval cue; and (2) the required information must
have been stored in main memory. In the case of dialogue C,
either requirement (1) or (2) may not be satisfied.

The differences between the two models are summarized below:

\begin{itemize}

\item New intention subordinate to current intention:
\begin{itemize}

\item Stack: Push new focus space

\item Cache: New entities retrieved to cache related to new
intention, old entities remain until displaced

\end{itemize}

\item Completion of intention agreed by conversants explicitly
or implicitly \begin{itemize}

\item Stack: Pop focus space for intention from stack, entities
in focus space are no longer accessible

\item Cache: Don't retain entities for completed intention, but
they remain accessible by virtue of being in the cache until
they are displaced

\end{itemize}

\item New intentions subordinate to prior intention
\begin{itemize}

\item Stack: Pop focus spaces for intervening segments, focus
space for prior intention accessible after pop

\item Cache: Entities related to prior intention must be
retrieved from main memory to cache, unless  retained in the
cache

\end{itemize}

\item Returning from interruption

\begin{itemize}

\item Stack: Length and depth of interruption  and the
processing required is irrelevant

\item Cache: Length of interruption  or the processing required
predicts retrievals from main memory

\end{itemize}

\item Centering

\begin{itemize}

\item Stack: No clear relationship between the focus stack
mechanism and centering \cite{GS85}; (Grosz and Sidner, this
volume)

\item Cache: Centers are a subset of the elements in the cache
and centering provides a finer level of salience ordering for
entities in the cache.

\end{itemize}

\end{itemize}

In the next section, I describe how the centering algorithm is
integrated with the cache model.

\section{Integrating the Centering Algorithm with the Cache
Model} \label{alg-sec}

Brennan, Friedman and Pollard (1987) proposed a centering algorithm
for the resolution of third person anaphors, based on the centering
rules and constraints, whose top level structure is shown in Figure
2. This section presents a version of that algorithm that is
integrated with the cache model by assuming that the Cf list is a
subset of entities available in the cache.  The revised algorithm also
incorporates observations from a corpus analysis of centering
\cite{Walker89b}, experimental processing results
\cite{NicolSwinney89,GMR92,Brennan95,Hudson88,GGG93}, and proposals in
Brennan {\it etal.}  of simple ways to make the algorithm more
efficient. This section extends and integrates work previously
presented in \cite{BFP87,Walker89b,WIC90,WIC94,Walker96b}.

\begin{figure}[htb] \begin{small} \begin{center}
\rule{14cm}{.2mm} {\bf CENTERING ALGORITHM} \end{center}
\begin{enumerate} \item CONSTRUCT THE PROPOSED ANCHORS for U$_n$
\item INTERLEAVE CREATION AND FILTERING OF PROPOSED ANCHORS
\item  UPDATE CONTEXT    \end{enumerate} \rule{14cm}{.2mm}
\end{small} \label{top-cent-alg-fig} \caption{ Top Level
Structure of the Centering Algorithm (Brennan, Friedman and
Pollard, 1987)} \end{figure}
\begin{figure}[htb] \begin{small} \begin{center}
\rule{14cm}{.2mm} {\bf CONSTRUCT THE PROPOSED ANCHORS for U$_n$}
\end{center}

\begin{enumerate}

	   \item Create set of referring expressions (REs). REs
represent discourse entities in the representation of the
discourse model.  If there is a conjoined NP, make one RE whose
extension is both entities. \footnote{ \cite{Birner97} (this
volume) suggests that the poset relation may be treated as a
simple inference for subsequent reference.}

   \item Order REs by the Cf ranking for the language. Cf
rankings are typically derived from a combination of syntactic,
semantic and discourse features associated with entities evoked
by the utterances in a discourse.

   \item Create set of possible forward center (Cf) lists. 
Expand each element of (b) according to whether it is a pronoun,
a description, or a proper name.  These expansions are a way of
encoding a disjunction of possibilities.

	\begin{enumerate} 	\item Expand pronouns into set with entry
for each RE in the Cf(U$_{n-1}$) that is consistent with: \\

(1) its agreement features; \\ (2) the selectional constraints
projected by the verb; \\ (3) the contraindexing constraints of
other elements in the current Cf list being expanded. \\ If
pronouns cannot be expanded by unification with entities in
Cf(U$_{n-1}$), then goto 4.

	\item Descriptions are not expanded, rather they are

represented by their intension and an index. Goto 5.

	\item Expand proper nouns into a set with an entry 	for each
discourse entity it could realize. Goto 5.

	\end{enumerate}

\item  First, attempt to expand pronouns by unification with
entities in the cache. If this returns null, reinstantiate the
contents of the cache by using the pronominal features and the
content of the utterance as retrieval cues for retrieving
matching discourse entities from main memory. Then goto 5.

   \item Create list of possible backward centers (Cbs).  This
is  the REs from step 3 or 4  plus an additional entry of NIL to
allow the possibility that the current utterance has no Cb.

\end{enumerate} \end{small} \rule{14cm}{.2mm}
\label{construct-alg-fig} \caption{ First Step of the Centering
Algorithm} \end{figure}

The centering algorithm starts with a set of reference markers for each
utterance.  Reference markers are generated for each referring expression in
an utterance and are specified for agreement, grammatical function, and
selectional restrictions; the values for these attributes arise from the
verb's subcategorization frame
\cite{PollardSag87,Reinhart76,Dieugenio90,WIC94,Dieugenio97,Passonneau95}.\footnote{Neither
predicative noun phrases e.g. {\it a beauty} in {\it Justine was a beauty},
nor pleonastic NPs such as {\it it} in {\it It was raining} count as referring
expressions.} . Reference markers are also specified for contraindices, which
are pointers to other reference markers that a marker cannot co-specify with
\cite{Reinhart76,PollardSag87};\footnote{See \cite{Sidner83a} for definition
and discussion of co-specification.}  these are calculated during
parsing. Each pronominal reference marker has a unique index from $A_{1},
\ldots ,A_{n}$ which will be linked to the semantic representation of the
co-specifier.  For non-pronominal reference markers the surface string is used
as the index.  Indices for indefinites are generated from $X_{1}, \ldots
,X_{n}$.

The first step of the centering algorithm is given in Figure 3; substep 4 of
Figure 3 specifies how centering is integrated with the cache model. At the
end of Step 1, the algorithm returns a set of potential Cbs and Cfs. The
second step of the algorithm is given in Figure 4. Figure 4 specifies how
potential anchors (Cb-Cf combinations) are created, in the order of preference
according to centering transitions. These anchors are then filtered further by
Constraint 3 and Rule1 of the centering rules and constraints (Cf. Walker,
Joshi and Prince, this volume). The first anchor to pass all the filters is
used to update the context (Step 3 of the algorithm).

\begin{figure}[htb] \begin{small} \begin{center}
\rule{14cm}{.2mm} {\bf INTERLEAVE CREATION AND FILTERING OF
PROPOSED ANCHORS} \end{center}

   \begin{enumerate}

   \item Create the proposed anchors, the Cb-Cf combinations
from the cross-product of the previous two steps, in order of
preferred interpretations.  Apply filters to each created anchor
in order of preference.

\begin{enumerate} \label{gen-step} \item Create {\sc continue}
anchors. Go to \ref{filter-step}. \item Create {\sc retain}
anchors. Go to \ref{filter-step}. \item Create {\sc smooth
shift} anchors. Go to \ref{filter-step}. \item Create {\sc rough
shift} anchors. Go to \ref{filter-step}. \item Create {\sc Null
Cb} anchors. Go to \ref{filter-step}. \end{enumerate}

   \item For each anchor in the current list of anchors apply
the following filters derived from the centering constraints and
rules. The first anchor that passes each filter is used to
update the context.  If more than one anchor at the same ranking
passes all the filters, then the algorithm predicts that the
utterance is ambiguous.  \label{filter-step}

\begin{enumerate}   \item FILTER 1: Go through Cf(U$_{n-1}$)
keeping (in order) those which appear in the proposed Cf list of
the anchor.  If the proposed Cb of the anchor does not equal the
first element of this constructed list then eliminate this
anchor. This guarantees that the Cb will be the highest ranked
element of the Cf(U$_{n-1}$) realized in the current utterance. 
This corresponds to constraint 3.

      \item FILTER 2: If none of the entities realized as
pronouns in the proposed Cf list equals the proposed Cb then
eliminate this anchor. If there are no pronouns in the proposed
Cf list then the anchor passes this filter. This corresponds to
Rule 1 by guaranteeing that if any element is realized as a
pronoun then the Cb is realized as a pronoun. \item If the
anchor doesn't pass the filters then goto \ref{gen-step} and try
the anchors for the next lower ranked transition type. Otherwise
goto Step 3, UPDATE CONTEXT. \end{enumerate} \end{enumerate}
\end{small} \rule{14cm}{.2mm} \label{interleave-alg-fig}
\caption{ Second Step of the Centering Algorithm} \end{figure}

\begin{figure}[htb] \begin{small} \begin{center}
\rule{14cm}{.2mm} {\bf UPDATE CONTEXT} \end{center}   If one of
the anchors passes all the filters then choose that anchor for
the current utterance.  Set Cb(U$_{n}$) to the proposed Cb and
Cf(U$_{n}$) to proposed Cf of this anchor. This will be the most
highly ranked anchor. \end{small} \rule{14cm}{.2mm}
\label{update-alg-fig} \caption{ Third Step of the Centering
Algorithm} \end{figure}

The difference between the algorithm above and that in \cite{BFP87} is the
point at which the different filters are applied, the definition of where the
algorithm stops, and the integration with the cache model.\footnote{Filter 2
could be implemented as a preference strategy rather than a strict filter, and
the violation of this rule could generate an implicature \cite{GHZ93}, or
possibly function as a new segment indicator
\cite{Fox87,PassonneauLitman96}. See \cite{Nakatani93a,WP94,Cahn95} for a
discussion of the difference between accented and unaccented NPs in this
role.} In \cite{BFP87}, all potential anchors were generated and then
filtered.  Here fewer anchors are generated even in the worst case since some
filters apply to potential Cf lists before the anchors are generated. In
particular, filtering by contraindices is included earlier both for efficiency
and because there is experimental evidence that this constraint is applied
very early \cite{NicolSwinney89}. In addition, since the anchors are generated
in preference order and then filtered, many fewer anchors are typically
generated. For example in Dialogue D, a constructed monologue used by
\cite{BFP87} to illustrate the centering algorithm, only three anchors are
generated where the original algorithm generated sixteen.

\eenumsentence[D]{ \item[a.] Susan drives an Alfa Romeo. 

\item [b.]  She drives too fast. 

\item [c.] Lyn races her on weekends. 

\item[d.] She often beats her. 

}

Finally, the algorithm allows pronouns to be resolved to
entities in the cache whenever pronouns cannot be unified with
centers from the previous utterance.

\section{Evidence for the proposed integrated model}

Remember that centering was formulated  as a process that
operates on two utterance U$_n$ and U$_{n+1}$, {\it within a
discourse segment D}, which attempts to explain the relationship
between the form of referring expressions and underlying
discourse processes, While Grosz and Sidner, (this volume)
suggest that discourse segmentation affects the accessibility of
centers, the hypothesis considered here is that the
within-segment constraint should be abandoned. Furthermore, in
the proposed integrated model, the cache contents, rather than
discourse segment structure, determines the accessibility of
centers.

To support the proposed integrated model, this section presents
three types of evidence.  Section \ref{fs-sec} presents evidence
that the cache model can handle `focus pops', which were believed
to provide strong support for Grosz and Sidner's stack model. 
Then section \ref{now-sec} discuss quantitative evidence showing
that centers are frequently carried over segment boundaries.
Finally section \ref{seg-sec} discuss a number of naturally
occurring examples that illustrate that the form in which
centers are realized across discourse segment boundaries is not
determined by boundary type.

\subsection{Modeling Focus Pops with The Cache Model}

\label{fs-sec}

Sometimes in a discourse, the conversants return to the discussion of a prior
topic or continue an intention suspended in prior discourse. This kind of
return has given rise to a phenomenon called {\sc return pops} or {\sc focus
pops}, in reference to the stack mechanism which pops intervening focus spaces
\cite{PS84,Reichman85,GS86}. The phenomenon that characterizes {\sc return
pops} is the occurrence of a pronoun in an utterance, where the antecedent for
the pronoun is in the focus space representing the prior discourse, that is
hierarchically recent.  Thus it is commonly believed that this provides strong
motivation for the role of hierarchical recency, and thus for Grosz and
Sidner's stack model.

In the stack model, any of the focus spaces on the stack can be returned to,
and the antecedent for a pronoun can be in any of these focus spaces. As a
potential alternative to the stack model, the cache model appears to be unable
to handle return pops since a previous state of the cache can't be popped to.
Since return pops are a primary motivation for the stack model, I re-examine
all of the naturally-occurring return pops that I was able to find in the
literature
\cite{Grosz77,Sidner79,Reichman85,Fox87,PassonneauLitman96}.\footnote{Fox
provides some quantitative data on return pops with and without pronouns, that
show that return pops with pronouns in written texts are virtually nonexistent
\cite{Fox87}.} There are 21 of them. I argue that return pops are {\bf cued
retrieval from main memory}, that the cues reflect the context of the pop,
that the cues are used to reinstantiate the relevant cache contents, and thus,
that return pops are not problematic for the cache model.

As an example of a return pop, consider dialogue E
\cite{PassonneauLitman96}(figure 9):

\eenumsentence[E]{ \item[] 21.1 Three boys came out, \\ 21.2
helped him$_i$ pick himself up, \\ 21.3 pick up his$_i$ bike, \\
21.4 pick up the pears, \\ 21.5 one of them had a toy, \\ 21.6
which was like a clapper. \\ 22.1 And I don't know what you call
it except a paddle with a ball suspended on a string. \\ 23.1 So
you could hear him$_j$ playing with that. \\ 24.1 And then
he$_i$ rode off.  }  

In dialogue E, the sequence from 21.5 to
23.1 is an embedded segment. According to the cache model, the
cache is not automatically reset to contain the information from
the interrupted segment after the final utterance of an embedded
segment.  Thus either that information must be retained because
there is an expectation that it will be returned to, or at some
point after utterance 23.1, perhaps as a result of processing
24.1, the hearer must retrieve the necessary information from
main memory to the cache in order to reinstantiate it in the
cache and interpret the pronoun in 24.1.

In the cache model, there are at least three possibilities for
how the context is created so that pronouns in {\sc return pops}
can be interpreted: (1) The pronoun alone functions as a
retrieval cue \cite{GMR92}; or (2) The content of the first
utterance in a return indicates what information to retrieve
from mainmemory to the cache, which implies that the
interpretation of the pronoun is delayed; (3) The shared
knowledge of the conversants creates expectations that
determines what is in the cache, e.g.  shared knowledge of the
task structure. I leave this last possibility aside for now.

Let us consider the first possibility.  The view that pronouns
must be able to function as retrieval cues is contrary to the
classic view that pronouns indicate entities that are currently
salient, i.e.  in the hearer's consciousness
\cite{Chafe76,GHZ93,Prince81}.  However, there are certain cases
where a pronoun alone is a good retrieval cue, such as when only
one referent of a particular gender has been discussed in the
conversation. With {\sc competing antecedent} defined as one
that matches the gender and number of the pronoun \cite{Fox87},
Figure 6 shows the distribution of the 21 return pops found in
the literature according to whether competing antecedents for
the pronoun are elements of the discourse model.

\begin{figure}[htb] \begin{center} \begin{tabular}{|c|c|} \hline
&\\ Competing Referent & No Competing Referent   \\ \hline  11 &
10  \\ \hline  \end{tabular} \end{center} \caption{Number of
Pops with Potentially Ambiguous Pronouns} \label{return-pop-fig}
\end{figure}

While it would be premature to draw final conclusions from such
a small sample size, the numbers suggest that in about half the
cases we could expect the pronoun to function as an adequate
retrieval cue based on gender and number cues alone. In fact,
Sidner proposed that return pops might always have this property
with her {\sc stacked focus constraint}: {\it Since anaphors may
co-specify the focus or a potential focus, an anaphor which is
intended to co-specify a stacked focus must not be acceptable as
co-specifying either the focus or potential focus. If, for
example, the focus is a noun phrase which can be mentioned with
an {\it it} anaphor, then {\it it} cannot be used to co-specify
with a stacked focus.} \cite{Sidner79}, p. 88,89.

However, since representations (reference markers) for centers in the
centering algorithm include selections restrictions from the verb's
subcategorization frame, we might reasonably define {\sc competing antecedent}
to reflect the fact that the center's representation includes selectional
restrictions \cite{Dieugenio90,Levin93}; Di Eugenio (this volume).\footnote{In
fact in languages with zero pronouns like Japanese, all the information is
contained in the verb subcategorization frame \cite{Iida92,WIC94}.}
Furthermore, we expect that these selectional restrictions are used as
retrieval cues.

Of the eleven tokens with competing referents in figure
\ref{return-pop-fig}, five tokens have no competing referent if
selectional restrictions are also applied.  For example, in the
dialogues about the construction of a pump from
\cite{Deutsch74}, only some entities can be bolted, loosened, or
made to work. Furthermore, if a selectional constraint can arise
from the dialogue, then only 4 pronouns of the 21 return pops
have competing referents.  For example, the verb {\it ride} in
dialogue E eliminates other antecedents because only one of the
male discourse entities under discussion has, and has been
riding, a bike \cite{PassonneauLitman96}.\footnote{Fox proposes
that lexical repetition is used as a signal of where to pop to
\cite{Fox87}, pps.  31,54.} Thus in 17 cases, an adequate
retrieval cue is constructed from processing the pronoun and the
matrix verb \cite{Dieugenio90}.

The second hypothesis is that the content of the return utterance indicates
what information to retrieve from main memory to the cache. The occurrence of
{\sc informationally redundant utterances} (IRUs) is one way of doing this
\cite{Walker93c,Walker96b}. For example, in dialogue F
\cite{PassonneauLitman96}, utterances 4 to 8 constitute a separate segment,
and utterance 9, which is the beginning of a return pop, is also an IRU,
realizing the same propositional content as utterance 3.

\eenumsentence[F]
{ \item[] (1)  a-nd his bicycle hits a rock. \\  
(2) Because {\it he$_i$'s} looking at the girl. \\   
(3)  {\it ZERO-PRONOUN$_{i}$} falls over, \\
(4)   uh there's no conversation in this movie.  \\ 
(5)   There's sounds, \\  (6)   you know,  \\ 
(7)   like the birds and stuff,  \\  
(8)   but there.. the humans beings in it don't say anything.  \\  
(9)   {\it He$_{i}$} falls over, \\ 
(10) and then these three other little kids about his same age come walking by. 
}

IRUs at the locus of a return can: (1) reinstantiate required
information in the cache so that no retrieval is necessary; (2)
function as excellent retrieval cues for information from main
memory.  Figure \ref{return-pop-iru-fig} shows the distribution
of IRUs in the 21 return pops found in the literature. The fact
that IRUs occur in 6 cases shows that IRUs are often used to
recreate the relevant context. IRUs in combination with
selectional restrictions leave only 2 cases of pronouns in
return pops with competing antecedents.

\begin{figure}[htb] \begin{center} \begin{tabular}{|c|c|} \hline
&\\ with IRU & without IRU \\ \hline 6 & 15 \\ \hline
\end{tabular} \end{center} \caption{Number of Pops with Pronouns
with and without IRUs} \label{return-pop-iru-fig} \end{figure}

In the remaining 2 cases, the competing antecedent is not and
was never prominent in the discourse, i.e. it was never the
discourse center \cite{Iida97}. This lack of prominence suggests
that it may never compete with the other cospecifier.

Thus, while more evidence is needed, it is plausible that the
cache model can handle this well-known phenomenon, by positing
that a return pop is a {\bf cued retrieval from main memory} and
that return pops never occur without an adequate retrieval cue
for reinstantiating the required entities, properties and
relations in the cache.

\subsection{Distribution of Centering Transitions in Segment
Initial Utterances}

\label{now-sec}

\begin{figure}[htb] \begin{center} \begin{tabular}{|r|c|c|c|c|}
\hline &&&&\\ & Continue & Retain & Smooth-Shift & No Cb \\ 
\hline \hline &&&&\\  Segment initial {\it now} sentences & 2 &
20 & 38 & 38 \\ \hline Other Sentences & 43 & 9 & 27 & 21 \\
\hline \end{tabular} \end{center} \caption{Distribution of
Centering Transitions in 98 discourse-segment initial {\it Now}
sentences as compared with a control group of Other sentences
from (Hurewitz, 1995)}  \label{cent-distrib-fig} \end{figure}

 One way to see whether discourse segment structures have a
direct effect on centering data structures is by examining
differences in the centering transitions across discourse
segment boundaries, which indicates whether centers are carried
over utterance pairs that span discourse segment boundaries. 
The cache model predicts that centers are carried over segment
boundaries by default because they are elements of the cache,
but that the recognition of a new intention may have an effect
on centering because it may result in a retrieval of new
information to the cache. It also predicts that the degree to
which centers are carried over or retained depends directly on
whether they continue to be used in the new segment (because the
cache replacement policy is to replace the least recently
accessed (used) discourse entities). This means that discourse
segmentation should have a gradient effect on centering.

Figure \ref{cent-distrib-fig} shows centering transitions in 98
segment initial utterances \cite{Walker93f}, where discourse
segment boundaries were identified by the use of the cue word
{\it now} \cite{HL87}.\footnote{See Walker, Joshi and Prince
(this volume) for the definition of the centering transitions of
{\sc continue, retain, smooth-shift, rough-shift} and {\sc no
cb}. In the data here, no rough-shift transitions were found.} 
{\it Now} indicates a new segment that is a further development
of a topic, and indicates a push in the stack model
\cite{GS86,Reichman85,HirschbergLitman93}. To my understanding
this means that discourse segments that are initiated with
utterances marked by the cue word {\it now} are either sister
segments or subordinated segments.

The figure shows that centering transitions distribute
differently for this type of segment initial utterance than they
do for utterances in general. \footnote{I have taken the liberty
of converting Hurewitz's percentages to raw numbers based on a
sample of 100 tokens.} A similar distributional difference in
centering transitions is reported in \cite{Passonneau95}. The No
Cb cases in Figure 8 indicate that there are some new segments
where centers are {\bf not}  carried over, but note that even
within a discourse segment, centers may not be carried over from
one utterance to the next. In addition, in about two thirds of
the segment initial utterances,
centers {\bf are} carried over discourse segment boundaries, so
that there is a gradient effect of discourse segment boundaries
on centering.

These distributional facts demonstrate the need for a model of
global focus that is integrated with centering, and provides
support for the proposed cache model since centers are clearly
carried over segment boundaries, and since there is a gradient
effect of segmentation on centering transitions.

\subsection{Discourse Configurations and Centering Data
Structures} \label{seg-sec}

 This section presents data showing that discourse segment
structure does not determine the accessibility of centers.  It
is well known that accessibility of discourse entities is
reflected by linguistic form
\cite{GHZ93,Prince81,Prince92,Brennan95}. Furthermore,
psychological studies of centering have shown that a processing
penalty is associated with realizing the Cb by a full noun
phrase (Hudson, this volume), \cite{GGG93}. Thus below the
realization of the Cb (linguistic form) is used as an indicator
of whether discourse segmentation has a direct effect on
accessibility.  

\begin{figure}[htb]
\centerline{\psfig{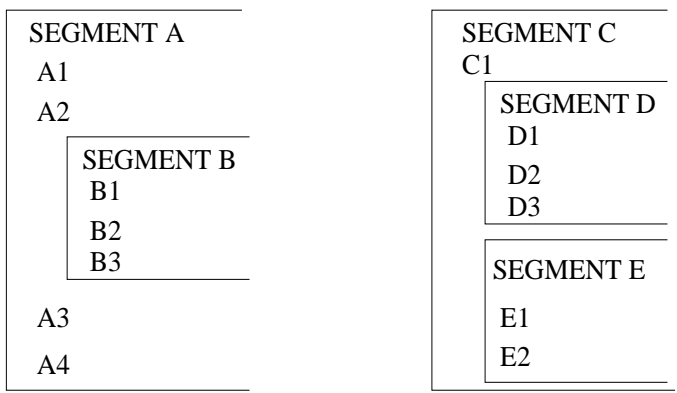}
} \caption{Two abstract hierarchical discourse structures. The
first has two discourse segments A and B where B is embedded
within A, and the second has three segments C, D, E where D and
E are sister segments contributing to the purpose of segment C. 
Utterances are represented as A1, A2 etc.} \label{hier-fig}
\end{figure} 

In order to show that discourse segment structure doesn't
determine accessibility, we must examine the linguistic form of
centers across all potential discourse segment structure
configurations. This means we must define all potential
discourse structure configurations.  Figure \ref{hier-fig}
illustrates different discourse structures in Grosz and Sidner's
theory and shows how segments consist of groupings of utterances
which can be embedded within one another. These discourse
structure configurations vary in terms of whether two utterances
can be considered to be linearly recent or  hierarchically
recent.

In Figure \ref{hier-fig}, utterance A1 is both linearly and
hierarchically recent for A2.  Since the utterances before and
after segment B are both part of segment A, utterance A2 is
hierarchically recent when A3 is interpreted, although it is not
linearly recent. Utterance B3 is linearly recent when A3 is
interpreted, but not hierarchically recent. Similarly B3 is not
hierarchically recent for A4.  In the second discourse, C1 is
hierarchically recent for both D1 and E1, but only linearly
recent for D1.  Utterance D3 is linearly recent for E1, but not
hierarchically recent.

Linear recency approximates what is in the cache because if
something has been recently discussed, it was recently in the
cache, and thus is is more likely to still be in the cache than
other items.  Linear recency ignores the effects of
preferentially retaining items in the cache, and retrieving
items from main memory to the cache. However linear recency is
more reliable as a coding category since it only relies on what
is indicated in surface structures in the discourse.

\begin{figure}[htb] \begin{small} \begin{center}
\begin{tabular}{|r|c|c|c|c|} \hline & & &&\\ Center &   Sister &
Subordinate & Focus Pop & Focus Pop \\ realization &  intention
& intention & Hierarchical, & Linear, \\ over U$_{n-1}$, U$_{n}$
& Over D3,E1&  Over C1,D1 & Over A2,A3& Over B3,A3 \\ & & & over
C1, E1 & \\ \hline \hline  &&&& \\ Cb $=$ PRONOUN & Type 1 &
Type 3 & Type 5 & Type 7 \\ \hline  Cb $=$ FULL NP & Type 2 &
Type 4 & Type 6 & Type 8 \\ \hline  \end{tabular} \end{center}
\caption{Centering and Discourse Segmentation Possibilities}
\label{cent-seg-fig} \end{small} \end{figure}

Given these terms, Figure \ref{cent-seg-fig} enumerates all the
relevant discourse structure configurations.  The columns of
Figure \ref{cent-seg-fig} are the types of discourse segment
boundaries that two utterances U$_{n-1}$ and U$_n$ can span in
terms of intentional structure and linear and hierarchical
recency. The rows enumerate differences in linguistic form that
are known to indicate center accessibility, i.e. whether the
Cb(U$_{n-1}$) is realized in U$_n$ as a pronoun or as a Full NP.
The combination of these two dimensions defines eight discourse
situations.

Types 1 and 2 are utterance pairs that are linearly recent but
not hierarchically recent because a related sister segment, e.g.
segment D, has already been popped off the stack. Types 3 and 4
are utterance pairs that are both linearly and hierarchically
recent.  Types 5 and 6 are utterance pairs where U$_{n-1}$ is
hierarchically recent but not linearly recent. Types 7 and 8 are
utterance pairs where U$_{n-1}$ is linearly recent but not
hierarchically recent, because an unrelated interrupting segment
has been popped off the stack.

To test the hypothesis that segment structure does not determine
accessibility, we must examine naturally occurring text or
dialogue excerpts that exemplify each configuration. See
Appendix A for a specification of criteria used to identify
relevant examples.  The remaining sections each discuss two of
the discourse types from Figure \ref{cent-seg-fig} using
excerpts from the Harry Gross corpus \cite{PHW82,Walker93c}, the
SwitchBoard Corpus from the LDC, Phil Cohen's corpus of
telephone-based dialogues between an expert and an apprentice
who must put together a plastic water pump \cite{Cohen84a}, and
excerpts from the Pear Stories Corpus from
\cite{PassonneauLitman93,PassonneauLitman96}. Centers are
indicated by italics and discourse segment structures are marked
by horizontal lines in the transcripts of the
discourse.\footnote{There may be additional segment structure
beyond what is indicated.}

\subsubsection{Type 1 and 2:  Sister intention} \label{type1-sec}

\begin{small} \begin{figure}[htb] \begin{center}
\begin{tabular}{|ccl|} \hline && \\ Seg$_i$ & U$_j$ & \\  \hline
\hline &&\\  6 & 28 & And you think ``Wow, \\  & & this little
boy's$_i$ probably going to come and see the pears,\\  & 29a&
and he$_i$'s going to take a pear or two,\\  &29b& and then go
on {\it his$_i$} way.''\\ \hline  &&\\  7 & 30 & um but {\it the
little boy$_i$} comes, ({\sc continue})\\  & 31& and uh he$_i$
doesn't want just a pear,\\  & 32& {\it he$_i$} wants a whole
basket.\\ \hline &&\\  8 & 33 & So {\it he$_i$} puts the bicycle
down,  ({\sc continue})\\  & 34 & and he$_i$ ..  you wonder how
he$_i$'s going to take it with this. \\ \hline  \end{tabular}
\label{pass-lit-fig} \caption{Excerpt from (Passonneau and
Litman, 1994) illustrating Type 1 and Type 2. Each line
indicates an empirically verified discourse segment.}
\end{center}

\end{figure} \end{small}

A sister intention discourse configuration is shown in Figure
\ref{hier-fig} for segments D and E; E is a sister to D.  The
Pear Stories narrative in Figure 11 from
\cite{PassonneauLitman96} illustrates two sister intention
discourse segments,\footnote{Based on assumption 4 (Appendix A),
segment 7 is a sister of segment 6 and segment 8 is a sister of
segment 7. } with the segment boundaries marked between
utterances 29 and 30 and between utterances 32 and
33.\footnote{These boundaries are those marked by a significant
number of naive subjects in Passonneau and Litman's experiments.}

 Consider the segment boundary spanned by utterance 29b and
utterance 30.  In segment 7, utterance 30, the full noun phrase
{\it the little boy} realizes the Cb of utterance 30
\cite{Passonneau95}. The discourse entity for {\it the little
boy} is also the Cb of utterance 29b and the Cp of utterance 30,
so the centering transition is a {\sc continue}.  Thus, this is
an example of Type 2 in Figure \ref{cent-seg-fig}: the
Cb(U$_{n-1}$) is realized as a full NP across a segment boundary
for two sister segments.

Now, consider the relation between utterance 32 and utterance 33
spanning the second segment boundary. Utterance 33 is also
segment initial, and the discourse entity for {\it the little
boy} is the Cb, but in this case this entity is cospecified by
the referring expression {\it he}. Here, as in utterance 30, the
discourse entity for {\it the little boy} is the Cb of the
previous utterance, utterance 32, and the Cp of the current
utterance, utterance 33, defining a {\sc continue} transition.
Thus, this is an example of Type 1 in Figure \ref{cent-seg-fig}:
the Cb of 32 is realized as a pronoun across a segment boundary.

Clearly, both Type 1 and Type 2 {\bf can} occur and  the Cb of
an utterance can be continued by means of a pronoun in the
initial utterance of a sister segment. Because a pronoun can be
used in this configuration, there is little motivation for
introducing an additional mechanism besides
centering to explain the accessibility of the center over sister
segment boundaries. The use
of the pronoun here can be explained quite naturally by assuming
that centering operates
over sister segment utterances, represented abstractly in Figure
9 by D3 and E1.

\subsubsection{Type 3 and 4: New subordinated intention}

\begin{small} \begin{figure}[htb] \begin{tabular}{|cccl|} \hline
&& & \\ Seg$_i$ & U$_j$ & Speaker$_k$ &\\  \hline \hline && & \\
N &(32) & H:& If you'd like a copy of that little article \\ &&&
just send me a note. \\ &&&I only have one copy. \\ &&& I'd be
glad to send {\it it} to you.\\ \hline && & \\ N+1 &(33) & C:&
Where did {\it it} appear? ({\sc continue})\\ &(34) & H:& it- I
- to tell you the truth\\ &(35) & C:& It wasn't in the newsp--\\
&(36) & H:& I don't remember where, \\ &&& what publication it
was. \\ &&& It was not a generally public thing like a
newspaper... \\ \hline  \end{tabular} \caption{Excerpt from the
Financial Advice Corpus illustrating Type 3. The discourse
segmentation is based on assumptions about the structure of
clarifications \protect\cite{Litman85}.} \label{type3-fig}
\end{figure} \end{small}

A new subordinated intention defines a new discourse segment embedded within
the immediately preceding segment, as segment D is embedded within segment C
in Figure \ref{hier-fig}.  Figure 12 consists of an excerpt from the financial
advice dialogue corpus \cite{PHW82}, showing one segment boundary. This
segment boundary is based on the assumption that a clarifying question
initiates a new discourse segment \cite{Litman85}.  Utterance 33 is a segment
initial utterance that refers to the Cb of utterance 32 with the referring
expression {\it it}.\footnote{Modulo the assumption that the article and a
copy of the article are being treated as coreferential.} Since this is the
only center on the Cf, it is also the Cb, resulting in a {\sc continue}
centering transition.  Thus, Figure 6 is an example of Type 3: utterance 33
shows that a Cb can be continued with a pronoun across a segment boundary
where the second segment is embedded within the first.

\begin{small} \begin{figure}[htb] \begin{tabular}{|ccrl|} \hline
&& & \\ Seg$_i$ & U$_j$ & Speaker$_k$ &\\  \hline \hline && & \\
N & 1 & Expert: & Now take the blue cap with the two prongs
sticking out \\ & 2  &Expert: & and fit the little piece of pink
plastic on {\it it}. Ok?   \\ & 3  &Apprentice: & Ok.  \\ \hline
&& & \\ N+1 & 4  &Expert: & Insert the rubber ring into {\it
that blue cap}. ({\sc retain}) \\ \hline  \end{tabular}
\caption{Excerpt from Pump Dialogue Corpus (Cohen, 1984)
illustrating Type 4. The discourse segmentation is based on the 
task structure (Grosz, 1977;Sibun,1991).} \label{pump-dial-fig}
\label{type4-fig} \end{figure} \end{small}

Figure 13 is an excerpt from Cohen's corpus of task-related
dialogues about the construction of a toy water pump
\cite{Cohen84a}, with one segment boundary indicated.  Here, the
segment boundary is based on the assumption that a new subtask
initiates a  subordinated segment \cite{Grosz77}.\footnote{In
this part of the dialogue, the goal is to put the blue cap and
its subcomponents onto the main pump body.  The rubber ring is a
subcomponent of the blue cap.  Thus putting the rubber ring into
the blue cap is a subgoal of adding the blue cap to the main
pump body.} This is an example of Type 4 because the Cb of
utterance 3 is cospecified by a deictic NP, {\it that blue cap},
in utterance 4. In this case, the previous Cb is not predicted
to be the Cb of the following utterance since the centering
transition is a {\sc retain}, and this
may be one factor involved in the choice of a deictic NP for the
referring expression.

Clearly both Type 3 and Type 4 {\bf can} occur, These types realize utterance
pairs that are both linearly and hierarchically recent, and show that the Cb
of the initial utterance of a subordinated segment can be expressed with
either a full NP or a pronoun. Thus, it is plausible that centering operates
over segment boundaries for subordinated segments, represented abstractly by
the relation between C1 and D1 in Figure 9.

\subsubsection{Type 5 and 6: Focus Pop with Hierarchical
Recency} \label{type56-sec}

 \begin{small} \begin{figure}[htb] \begin{tabular}{|ccl|} \hline
&& \\ Seg$_i$ & U$_j$ & \\  \hline \hline &&  \\ 14 & 1 & a-nd
his bicycle hits a rock. \\ & 2 & Because {\it he$_{i}$'s}
looking at the girl. \\  & 3 & {\it ZERO-PRONOUN$_{i}$} falls
over, \\ \hline && \\ 15 & 4 &   uh there's no conversation in
this movie.  \\ & 5 &  There's sounds, \\  & 6 &  you know,  \\
& 7 &  like the birds and stuff,  \\ & 8 &  but there.. the
humans beings in it don't say anything.  \\ \hline && \\ 16 & 9
&   {\it He$_{i}$} falls over,  \\ & 10 &  and then these three
other little kids about his \\ & & same age come walking by. \\
\hline  \end{tabular} \caption{An excerpt from the Pear Corpus
illustrating Type 5. Segment boundaries from human judgements
taken from Passonneau and Litman, 1994} \label{type5-fig}
\end{figure} \end{small}

In section \ref{fs-sec} we discussed focus pops, and argued that
focus pops could be
modeled with the cache model.   Here we are interested in
determining whether  the relevant structures for centering are 
determined by  hierarchical 
recency or by  linear recency of adjacent utterances. Thus, 
when a focus pop occurs there are two logical choices for
selecting U$_{n-1}$ for the purposes of centering, one choice
defined by linear recency and the other defined by hierarchical
recency. Types 5 and 6  select U$_{n-1}$ by hierarchical
recency.  In Figure \ref{hier-fig}, the relevant examples of
hierarchically recent utterances defined by focus pops let A2 be
U$_{n-1}$ for A3 and let C1 be U$_{n-1}$ for E1.

Figure \ref{type5-fig} is from the Pear Stories corpus, with
discourse segment boundaries marked by human judges
\cite{PassonneauLitman96}. This is a naturally occurring
exemplar of the first discourse in Figure \ref{hier-fig};
segment 15 is an interruption and segment 16 is a continuation
of segment 14. This analysis is also supported by: (1) the
obvious change in content and lexical selection
\cite{MorrisHirst91,Hearst94}; and (2) the fact that utterance 9
is an {\sc informationally redundant utterance}, IRU, which
re-realizes the content of utterance 3, and reintroduces its
content in the current context \cite{Walker93c,Walker96b}. Thus,
using hierarchical recency to determine U$_{n-1}$ for the
purposes of centering, U$_n$ is utterance 9 at the beginning of
segment 16 and U$_{n-1}$ is utterance 3 at the end of segment
14. Then, Figure \ref{type5-fig} is an example of Type 5 because
a pronoun is used in utterance 9 to realize the Cb of utterance 
3, despite the intervening segment 15.

Figure \ref{type6-fig} is an excerpt from the Switchboard corpus
in which the topic of the discussion was {\it Family Life}. The
discourse segment boundaries shown here were identified on the
basis of the claim that the cue word {\it anyway} marks a focus
stack pop to an earlier segment \cite{PS84,GS86,Reichman85}.
Utterance 5 in segment 3 starts with the cue word {\it anyway}
and returns to the discussion of which sports the speaker's
oldest son likes, after a brief digression about the speaker's
little girl.  Figure \ref{type6-fig} is an example of Type 6
because this focus pop realizes the Cb of utterance 3 with a
full NP, {\it my oldest son}. Note that no other male entity has
been introduced into the conversation, so on the basis of
informational adequacy alone, the pronoun {\it he} would have
sufficed \cite{Passonneau96}.

\begin{small} \begin{figure}[htb] \begin{tabular}{|ccrl|} \hline
& && \\ Seg$_i$ & U$_j$ & Speaker$_k$ &\\  \hline \hline && & \\
1 & 1 & A: & What are some of the things that you do with
them?\\  & 2 & B: &   Well, my oldest son is eleven, \\ & 3 & 
&and {\it he} is really into sports.\\  \hline & && \\ 2 & 4 & &
And my little girl just started sports. \\  \hline & && \\ 3 & 5
& &Anyway, {\it my oldest son}, he plays baseball right now, \\ 
& 6 & & and he's a pitcher on his team, \\  & 7 && and he's
doing really well.\\    \hline  \end{tabular} \caption{An
excerpt from the Switchboard corpus illustrating Type 6. The
topic of discussion was Family Life., Segment boundaries based
on the cue word {\it anyway}} \label{type6-fig} \end{figure}
\end{small}

Types 5 and 6 are utterance pairs where U$_{n-1}$ is
hierarchically recent, but not linearly recent.  The existence
of Types 5 and 6 shows that the Cb of an utterance in a prior
discourse segment (A2) can be referred to by either a pronoun or
a full NP in the initial utterance of a return (A3). Since both
Type 5 and Type 6 {\bf can} occur, it would seem that popping
alone does not make strong predictions about the realization of
the Cb.

 \subsubsection{Type 7 and 8: Focus Pop with Linear Recency}

\begin{small} \begin{figure}[htb] \begin{tabular}{|ccrl|} \hline
& && \\ Seg$_i$ & U$_j$ & Speaker$_k$ &\\  \hline \hline && & \\
1 & 1 & A: &   Well, what do you know about Latin American
policies? \\

& 2 & B:  &    Well, I think they're kind of ambivalent,
really.......\\

&& & (23 intervening utterances about US support etc) \\

& 25a & A: &   Yep, that's about the lump sum of it.  \\ \hline
& && \\ 2  &25b &   &   Well, um, I was speaking with a, a woman
from, \\ &&& I believe she was from the Honduras or Guatemala,
\\ &&& or somewhere in there, \\ &25c&   & No, she was from El
Salvador --\\

& 26 & B:  &   Yeah.\\

& 27a & A: &   -- and, uh, she was from a relatively wealthy
family, \\ &27b&   & and when, uh, the Contras came into power,
of course with, uh,\\ \hline & && \\ 3 & 27c&     & oh, gosh
darn, what's his face, \\ &&& he's in, in Florida jail now,
Marcos --\\

& 28 & B:  &   Yeah, yeah.\\

& 29 & A: &   -- uh, no, he's, Marcos is Philippines,\\

& 30 & B:  &   Yeah, um, well, I'm blank [laughter] on it.\\

& 31 & A: &   Well, you know who I'm talking about.\\

& 32 & B:  &   I can see {\it his} face (( )) forget his name
[laughter].\\

& 33a & A: &   Yeah, I, I know it, uh, \\ \hline & && \\ 4  &
33b&  & Anyway, when {\it he} came into power, \\ &&& he 
basically just took everybody's property, you know, \\ &&      &
just assigned it to himself.\\

& 34 & B:  &   Yeah, kind of nationalized it --\\ \hline
\end{tabular} 
\caption{An excerpt from the Switchboard corpus
illustrating Type 7. The topic of discussion was Latin America.
Segment boundaries are identified based on the cue word {\it
anyway}, {\sc informationally redundant utterances}, and tense
changes from simple past to present.} 
\label{type7-fig}
\end{figure} \end{small}

In section \ref{type56-sec}, we examined focus pops where
U$_{n-1}$ for the purposes of centering was defined by
hierarchical recency.  In Types 7 and 8, utterance U$_{n-1}$ for
the purpose of centering is defined by linear recency, where
U$_{n-1}$ belongs to a segment that is popped off the stack
before, or at the time that, U$_n$ is processed.  The linearly
recent utterance is analagous to letting B3 be U$_{n-1}$ for A3
in Figure \ref{hier-fig}.

 The segment structures for both Figures \ref{type7-fig} and
\ref{type8-fig}, illustrating Types 7 and 8, are defined on the
basis that the cue word {\it anyway} marks a pop to a previous
discourse segment, as posited by \cite{Reichman85,GS86}. Thus in
Figure \ref{type7-fig}, utterance 33b begins a new segment and
in Figure \ref{type8-fig}, utterance 7 begins a new segment.
However, in order to examine the effect of hierarchical recency,
the beginning of the intervening segment that is to be popped
must be identified. In Figure \ref{type7-fig}, utterance 27a in
segment 2 is assumed to be hierarchically recent for utterance
33b in segment 4 based on the IRU {\it when X came into power}
in utterance 33b \cite{Walker93c}. In Figure \ref{type8-fig},
the tense change from past to past imperfect between utterances
3a and 3b is used to identify a discourse segment boundary
\cite{Webber88a}, so that segment 3 is hierarchically adjacent
to segment 1.

Figure \ref{type7-fig} shows a conversation from the SwitchBoard
corpus in which two subjects are discussing the topic {\it Latin
America}, as seen in A's conversational opener in utterance 1.
The segment boundary of interest is that between utterance 33a
and 33b. Segment 3, from utterances 27c to 33a, is about trying
to remember the name of the leader of the Contras, and
establishes centers for both the Contra leader and for the
discourse entity representing his name. Establishing his name is
a minimal part of the story that speaker A is trying to tell. 
Segment 4 continues the Cb of the Contra leader, and continues
the story begun in utterance 27a, as shown by the paraphrase of
{\it When the contras came into power with (the Contra leader)}.
Clearly segment 4 continues the intention initiated in utterance
27b. Thus the focus space stack for segment 3 should be popped
from the stack by the use of the cue word {\it anyway}. However,
the use of the pronoun {\it he} to refer to the Contra leader in
33b would not be supported by the focus space for segment 2 that
would be on the top of the stack after the pop, since segment 3
actually established this discourse entity as a center.


Figure \ref{type8-fig} is also an excerpt from the Switchboard
corpus. In this case, the topic of the discussion was {\it home
decorating}. In utterance 7, speaker A marks a focus pop with
the cue word {\it anyway}. But what intention is segment 3
related to? I identified utterance 3a as the last utterance of
the hierarchically adjacent segment because the past imperfect
tense is used in utterance 3b when the simple past was used for
utterance 3a \cite{Webber88b}.  In addition, it is plausible
that on semantic grounds segment 2 provides background for
segment 3 \cite{Hobbs85a}, and thus that the intention of
segment 2 must be realized before that of segment
3.\footnote{Thus it may satisfaction-precede it in the
terminology of \cite{GS86}.} Then, this is an example of Type 8
because the phrase {\it that color} in utterance 7 refers to the
Cb of utterance 5 from segment 2, when the focus space for
segment 2 should be popped off the stack.

\begin{small} \begin{figure}[htb] \begin{tabular}{|ccrl|} \hline
& && \\ Seg$_i$ & U$_j$ & Speaker$_k$ &\\  1 & 1 & A: &   Well,
I was just looking around my house \\ &&&  and thinking about
the painting that I've done. \\

& 2 & B:  &   Uh-huh.\\

& 3a & A: &   And the last time that, um, we tackled it, I did
the kitchen.\\  \hline &&&\\ 2 & 3b   & &  And I had gone
through a period of depression at one time \\   &  & &and
painted everything a dark, it was called a sassafras, \\   &  &
&it was kind of an orangish brown.\\

& 4 & B:  &   Okay.\\

& 5 & A: &   {\it It} was not real pretty.\\

& 6 & B:  &   Yeah.\\  \hline &&&\\ 3 & 7 & A: &   Anyway, so
the kitchen was one of the rooms that got hit \\ & & &  with
{\it that color}.\\

& 8 & B:  &   Uh-huh, I see.\\

& 9 & A: &   [Laughter] So I tried to cover it with white....\\
\hline  \end{tabular} \caption{An excerpt from the Switchboard
Corpus illustrating Type 8. The discussion topic was home
decorating. Segment boundaries identified by the use of the cue
word {\it anyway} and tense changes.} \label{type8-fig}
\end{figure} \end{small}

Types 7 and 8 are utterance pairs where U$_{n-1}$ is linearly recent but not
hierarchically recent, because the interrupting segment has been popped off
the stack.  The existence of Types 7 and 8 illustrate that the Cb of an
utterance in a 'popped' segment (B3) can be referred to by either a pronoun or
a full NP in the initial utterance of a new 'pushed' segment (A3).  Since both
Type 7 and Type 8 {\bf can} occur, there seems to be no basis for assuming
that the centering data structures are directly affected by popping to a prior
focus space on the stack. The occurrence of Type 7 is strong support for the
cache model since there is clearly a change of intention between utterances
33a and 33b, but centers as part of attentional state are carried over and
realized with pronominal forms that clearly indicate their accessibility.

\section{Discussion} 

\label{conc-sec}

Centering is formulated as {\it a theory that relates focus of
attention, choice of referring expression, and perceived
coherence of utterances, within a discourse segment}
\cite{GJW95}, p. 204.  In this chapter I argue that the
within-segment restriction of centering must be abandoned in
order to integrate centering with a model of global discourse
structure. I have discussed several problems that this
restriction causes.  The first problem is that centers are often
continued over discourse segment boundaries with pronominal
referring expressions whose form  is identical to those that
occur within a discourse segment.  The second problem is that
recent work has shown that listeners perceive segment boundaries
at various levels of granularity and that segment
boundaries are often fuzzy. If centering models a universal
processing phenomenon, it seems implausible that each listener's
centering algorithm differs according to whether they perceived
a segment boundary or not, especially as there is evidence that
centering is a fairly automatic
process (Hudson-D'Zmura and Tanenhaus, this volume).  The third
issue is that even for utterances within a discourse segment,
there are strong contrasts between utterances that are adjacent
within a segment because they are hierarchically recent and
utterances that are adjacent
within a segment and also linearly recent.

This chapter argues that an integrated model of centering and
global focus can be defined
that eliminates  these problems by replacing Grosz and Sidner's
stack model of attentional state, with an alternate model, the
cache model \cite{Walker96b,Walker93c}. In the cache model,
centering applies to discourse entities in the cache, and the
contents of the cache can be affected by the recognition of
intention. However centers are carried over segment boundaries
by default, and are only displaced from the cache when they are
not being accessed. When a digression requires the use of all
the cache, a return requires a retrieval from main memory to
reinstantiate relevant discourse entities in the cache. Since
this retrieval has some processing costs, the cache model
predicts a role for linear recency which is not predicted by the
stack model. The proposed model integrates centering with
discourse structure defined by relations between speaker
intentions.

To provide support for the proposed integrated model, I first
show, in Section \ref{alg-sec},  how the centering algorithm is
easily integrated with the cache model. Then,  in sections
\ref{fs-sec}, \ref{now-sec} and \ref{seg-sec}, I provide three
types of data that support the integrated model. First, I show
that 'focus pops' can be handled by the cache model by positing
that they correspond to cued retrieval from main memory. I show
how features of the utterance in which the focus pop occurs
provide information that functions  as an adequate retrieval cue
from main memory.

Second, I examine the distribution of centering transitions in
98 segment initial utterances.  I show that that centering
transitions distribute differently in segment initial
utterances, and in particular that {\sc continue} transitions
are less frequent.  However it is clear that centers are carried
over segment boundaries, as the cache model would predict.

Third, section \ref{seg-sec} examines every type of discourse 
structure configuration in order to explore the relationship
between centering and hierarchical intentional structure.  The
data suggests that intentional structure does not define a rule
that directly predicts whether a discourse entity will be
realized as a full NP or as a pronoun across a segment boundary.
Figure \ref{type7-fig} shows that even segments that have been
popped from the stack can provide a center across a discourse
segment boundary.

These findings provide support for the proposed cache model. 
Since centers are in the cache, they are carried over segment
boundaries by default. In contrast, in the stack model, the
focus space where the center was established has been popped off
the stack. The cache model predicts a statistical correlation
between intentional structure and changes in intentional state,
which would arise because a change of intention can trigger a
retrieval of information to the cache, as in the case of `focus
pops'. But in order for hearers to retrieve the correct
information to the cache, either automatically or strategically,
the utterance must provide an adequate retrieval cue
\cite{RatcliffMcKoon88,McKoonRatcliff92}.

The cache model is also consistent with results of other work,
and with psychological models of human working memory
\cite{Baddeley86}. For example Davis and Hirschberg proposed
that rules for synthesizing directions in text-to-speech must
treat popped entities as accessible and de-accent them
\cite{DavisHirschberg88}. Huang proposed that rules for the form
of referring expressions in argumentative texts must treat the
conclusions of popped sisters as salient \cite{Huang94}. Walker
argued that the cache model explains the occurrence of {\sc
informationally redundant utterances}, IRUs, such as utterance 9
in Figure \ref{type5-fig}, as a way of providing an adequate
retrieval cue for reinstantiating relevant information in the
cache \cite{Walker96b}.

However a number of open issues remain.  First, 
while previous work has shown that a processing penalty is
associated with the use of a full NP to continue the current Cb
\cite{Hudson88,GGG93}; (Hudson-D'Zmura and Tanenhaus, this
volume), a full NP is used to continue the Cb in the examples of
Types 2 and 6 (Figures 5 and 9).\footnote{In the other cases a
full NP is used in a {\sc retain} transition.}  Why does this
occur?  

One possibility is that the use of the Full NP is one of a
number of potentially redundant cues that the speaker has
available for signalling intentional structure, so that the
choice of a Full NP or a pronoun is not determined by the
current attentional state \cite{Fox87,Yeh95,Passonneau96}.

A second possibility is the Full NP is used to signal the
rhetorical relation of contrast
\cite{Fox87,MannThompson87,Hobbs85a}.  This would explain the use
of a Full NP for both Type 2 (Figure 5) and Type 6 (Figure 9),
and unify these two cases with observations by \cite{Fox87} and
by Di Eugenio (this volume).  In Figure 5, a contrastive
relation between utterances 29 and 30 is indicated by {\it but}.
These segments contrast with one another by presenting alternate
possible worlds of what {\bf might} have happened with what {\bf
did} happen. In Figure 9, the NP {\it my oldest son} is an
example of Left-Dislocation \cite{Prince85}, i.e the discourse
entity realized as {\it my oldest son} is realized in an initial
phrase, and then again by the pronoun {\it he} in subject
position.  One function of Left-Dislocation is to mark an entity
as already evoked in the discourse or in a salient set relation
to something evoked, and contrast is inferred from the marking
of a salient set relation \cite{Prince86}.  Note that if
contrast is determining the use of the full NP, we expect
overspecified NPs to occur just as frequently within discourse
segments as in segment initial utterances.

Finally, future work should investigate what constitutes an
adequate retrieval cue for focus pops and how a speaker's
choices about the forms of referring expressions interacts with
other retrieval cues, such as propositional information. In
order to do this, it would be useful to have a large corpus of
data tagged for intentional structure.

A larger tagged corpus would also allow us to go beyond the study here, which
simply showed that intentional structures do not appear to define a rule that
determines the accessibility of centers. More data on the {\bf frequency} with
which various forms of referring expressions are chosen in different
situations would be useful. \cite{WW90} showed that in mixed-initiative
dialogues, pronominal forms were more likely to cross discourse segment
boundaries when one speaker interrupted the other than when transitions
between segments were negotiated between the conversants.  \cite{Passonneau95}
discusses the frequency with which Full NPs are used to realize entities
currently salient in the discourse. In \cite{Walker93f}, the frequency of
various forms of referring expressions was calculated for the segment
boundaries discussed in section \ref{now-sec}.  (Brennan, this volume), shows
that speakers are about twice as likely to use a full NP rather than a pronoun
if an utterance intervenes between the pronoun and its antecedent in the
discourse, and that pronouns and full NPs are equally likely in the same
situation when there is no intervening utterance. More data of this type would
be useful in defining algorithms for the generation of referring expressions,
and for determining additional factors involved in the referring expression
choice.

 In conclusion, this chapter presents a model that integrates
centering with hierarchical discourse structures defined by
speaker intention. The important features of the proposed
integrated model are that it: (1) explains the differences in
felicity between Dialogues  B and C; (2) predicts that centers
are carried over discourse segment boundaries by default; (3)
predicts a gradient effect of discourse segment structure on
centering as we see in Figure \ref{cent-distrib-fig}; (4)
predicts that granularity of intention-based segmentation has no
effect on centering; (5) predicts an increase in processing load
for pronouns in focus pops; and (6) is consistent with
psychological models of human sentence and discourse processing.


\section{Acknowledgements}

This research was partially funded by ARO grant
DAAL03-89-C0031PRI and DARPA grant N00014-90-J-1863 at the
University of Pennsylvania and by Hewlett Packard, U.K.

I'd like to thank Steve Whittaker, Aravind Joshi, Ellen Prince,
Mark Liberman, Karen Sparck Jones, Bonnie Webber, Scott
Weinstein, Susan Brennan, Janet Cahn, Mitch Marcus, Cindie
McLemore, Owen Rambow, Barbara Di Eugenio, Candy Sidner, Ellen
Germain, Megan Moser, Christine Nakatani, Becky Passonneau, Pam
Jordan,  Jennifer Arnold and several anonymous reviewers for
extended discussion of the issues in this paper. 


\section{Appendix A}

In order to identify relevant examples in corpora of naturally
occurring discourses, the first difficulty is determining the
discourse segment structure of naturally occurring texts and
dialogues. This involves two separate issues:

\begin{enumerate} \item An algorithm is needed to divide running
speech into utterance units that are relevant to determining
centering transitions such as {\sc continue}. \item These
utterances must then be grouped into segment structures that
correspond to speaker intentions. \end{enumerate}

To address the first issue, as a working assumption, I adopt a
simple algorithm for dividing discourses into utterances,
loosely based on Hobbs' algorithm \cite{Hobbs76a}:

\begin{enumerate} \item  An utterance is  a clause with a finite
verb. \item  Each coordinated clause in a  complex sentence 
defines an utterance. The order of the utterances in the
discourse follows the order of the production of the conjuncts.
\item  The previous utterance for subordinated clauses is the
superordinate clause. \item  The Cf for a complex sentence with
subordinated clauses is the  Cf for the main clause, with the
Cfs of the subordinates appended. \item  An utterance following
a complex sentence with subordinated clauses takes the centering
data structures from the main clause of the complex sentence as
its input.  \item    Prompts such as {\it yeah, okay, uh huh} in
dialogue  (implicitly) realize the centers  from the previous
utterance. \end{enumerate}

This is consistent with findings from corpus-based work reported
in \cite{Walker89b,Kameyama86a,SuriMcCoy94}.  See
\cite{Hobbs76a,SuriMcCoy94,Passonneau94} (Kameyama, this volume)
for further discussion.

The second issue is producing a segmentation on the basis of
speaker intention or similar semantic categories. Determining
reliable ways to segment discourse is an active area of research
\cite{WS88,WW90,GroszHirschberg92,PassonneauLitman93,Hearst94,MoserMoore95,CarlettaIsard95,FlammiaZue95b}. However, in order to
identify examples that match the configurations, we do not need
a {\bf complete} segmentation of a discourse. Rather, what is
required is a method for identifying segment initial utterances
that stand in a particular configuration to utterances in prior
segments. Here, six criteria were used:

\begin{enumerate} \item The use of cue words such as {\it now}
and {\it anyway} are treated as reliable indicators of the
initiation of a discourse segment. Following the theories of
\cite{GS86,Reichman85,HirschbergLitman93} and empirical results
in \cite{Litman94}, both {\it now} and {\it anyway} indicate a
new segment. {\it Now} indicates a new segment that is a further
development of a topic, and indicates a push in the stack model.
{\it Anyway} is a cue to a return to a prior discussion, and
indicates a pop in the stack model. \item If the initiation of a
segment D1 is indicated by the  use of {\it anyway},  tense
changes and the occurrence of {\sc informationally redundant
utterances} (IRUs) are treated as indications of which prior
segment is related to the newly initiated segment (where to pop
to in the stack model) \cite{Webber88b,Walker93c}; \item 
Clarification questions are treated as initiators of new
subordinated discourse segments, following 
\cite{Sidner83b,Sidner85,Litman85,WW90,LambertCarberry91}; 
\item Discourse segments marked  by human judges on the Pear
Stories\footnote{This corpus consists of narrations of a movie
by a subject who had seen the movie to another subject who had
not seen the movie \cite{Chafe80}}, are taken from experiments
reported in
\cite{PassonneauLitman93,PassonneauLitman96,Passonneau95}. \item
All discourse segments in the Pear Stories are assumed to be
sister segments  on the basis that these narrations  relate a
temporal {\bf sequence} of events, and that if event A
temporally precedes event B, then the intention of segment A
must be realized before  the intention of segment B
\cite{Polanyi87,Webber88b,Sibun91}.  These event sequence
segments are dominated by the single intention of `telling the
story'.  

\item  In  Cohen's pump construction dialogues, if there is a 
goal and subgoal relationship between the content of the
segments and the structure of the task \cite{Grosz77,Sibun91},
then the subgoal segment is assumed to be embedded within the
goal segment. \end{enumerate}

This set of segment identification criteria is the basis for the
identification of naturally occurring discourses that fit in
each of the cells in Figure \ref{cent-seg-fig}.

\end{document}